# Spatial-Temporal Feature Extraction and Evaluation Network for Citywide Traffic Condition Prediction

Shilin Pu[1], Liang Chu[1], Zhuoran Hou[1], Jincheng Hu[2], Yanjun Huang[3], Yuanjian Zhang[2]

[1] College of Automotive Engineering, Jilin University, Changchun 130022, China.
[2] Department of Aeronautical and Automotive Engineering, Loughborough University, Loughborough, U.K.
[3] School of Automotive Studies, Tongji University, Shanghai 201804, China.
Corresponding Author: Yuanjian Zhang ( y.y.zhang@lboro.ac.uk )

**Abstract**: Traffic prediction plays an important role in the realization of traffic control and scheduling tasks in intelligent transportation systems. With the diversification of data sources, re asonably using rich traffic data to model the complex spatial-temporal dependence and nonlinear characteristics in traffic flow are the key challenge for intelligent transportation system. In addition, clearly evaluating the importance of spatial-temporal features extracted from different data becomes a challenge. A Double Layer - Spatial Temporal Feature Extraction and Evaluation (DL-STFEE) model is proposed. The lower layer of DL-STFEE is spatial-temporal feature extraction layer. The spatial and temporal features in traffic data are extracted by multi-graph graph convolution and attention mechanism, and different combinations of spatial and temporal features are generated. The upper layer of DL-STFEE is the spatial-temporal feature evaluation layer. Through the attention score matrix generated by the high-dimensional self-attention mechanism, the spatial-temporal features combinations are fused and evaluated, so as to get the impact of different combinations on prediction effect. Three sets of experiments are performed on actual traffic datasets to show that DL-STFEE can effectively capture the spatial-temporal features and evaluate the importance of different spatial-temporal feature combinations.

## I. INTRODUCTION

With the continuous acceleration of urbanization, the population and vehicle ownership are also increasing, resulting in traffic congestion and other problems. In order to improve the efficiency, sustainability and security of transportation network, intelligent transportation system (ITS) [1] is proposed and becomes an advancing research field. Traffic prediction is an important step in the development of intelligent transportation [2]. It aims to predict future traffic conditions by integrating historical observation data and measurement information of road sensor networks. A large number of sensors such as loop detectors, cameras and weather sensors are deployed and constantly generate a large number of traffic data, and provides support for the application of traffic prediction. At present, traffic forecasting is applied in many aspects, such as vehicle-environment cooperative control[3], passenger demand forecasting [4, 5], travel time estimation [6], order scheduling [7] and congestion control [8], etc.

Recently, the research methods of traffic prediction are mainly data-driven methods, which are divided into three categories: statistical methods, basic machine learning methods and deep learning methods. Statistical methods model the time series of traffic data, so as to learn the statistical characteristics of the data to predict the future traffic conditions. For example, ARIMA [9], Gaussian process [10, 11], hidden Markov model [12, 13], Bayesian network [14] and Kalman filter [15, 16]. However, these methods are largely limited by the assumption of time series stationarity. At the same time, they can only predict a single or a small number of traffic flow series at a time, and they are not easy to be extended to complex and large traffic data sets. More importantly, these methods cannot learn the nonlinear correlation in traffic data, that is, the complex spatial-temporal correlation characteristics. With the development of machine learning methods, basic machine learning methods are applied to the field of traffic prediction, such as k-nearest neighbor algorithm (KNN) [17], support vector machine (SVM) [18] and artificial neural network(ANN) [19, 20]. Compared with statistical methods, basic machine learning methods can model relatively more complex data and achieve higher prediction accuracy, but these methods also do not have the ability to model the complex nonlinear spatial-temporal correlation characteristics in traffic data, and it cannot automatically extract features from the input data. Therefore, feature engineering based on expert experience is an important part that affects the prediction accuracy.

Compared with basic machine learning methods, deep learning methods have a greater improvement in both the structural depth of the model and the ability to capture features. Specifically, the deep learning methods can more accurately capture the spatial-temporal correlation characteristics in traffic data, so it is widely used in traffic prediction tasks. At present, general deep learning models used in traffic prediction include convolutional neural network (CNN), recurrent neural network (RNN) and its variants, namely long short-term memory neural network and gate recurrent unit neural network (LSTM, GRU), graph convolution neural network (GCN), attention mechanism, etc.

The early researches are mostly based on single model to predict traffic flow. For the extraction of spatial correlation features of traffic data, CNN shows its advantages in the capture of European spatial correlation features and the complexity of the model [21]. However, the characteristics of traffic data are more expressed in the form of non-European space. GCN gains widespread attention thanks to its rich non-Euclidean spatial modeling features. Through different definition methods of non-European space, the multi graph convolution network can model the relevant characteristics of non-European space from different angles [22, 23]. For the extraction of nonlinear time-dependent features of traffic data, RNN and its variant LSTM/GRU benefit from the gating mechanism, showing advantages in capturing nonlinear time-dependent features in traffic data [24]. However, for long-term sequence RNN has the problem of long-term information forgetting, for long-span time feature capture, the attention mechanism has the ability to correlate data at any two moments, Therefore, it is commonly used to describe the periodic modeling task of traffic data [23].

With the improvement of the requirements for the accuracy of traffic prediction, the modeling of spatial or temporal correlation characteristics alone can no longer meet the requirements. Aiming at achieving a higher traffic prediction effect, the method of combining different deep learning models to capture the spatial-temporal correlation characteristics of traffic data at the same time has gradually become the mainstream. At the beginning of the research, the method of CNN and RNN model fusion was proposed [4, 25] to capture the correlation characteristics of European Space and time. With the development of GCN, a model fusion scheme based on GCN is proposed. For example, GCN combines with CNN model to extract the spatial-temporal characteristics of traffic data [26-28], or GCN combines with RNN model to extract spatial and temporal characteristics respectively [22, 29]. With the development of attention in the extraction of correlation features of sequence elements, a scheme using attention mechanism to extract the correlation characteristics of time and space [30, 31] has been proposed.

Although the integrated deep learning model shows great advantages and potential in traffic flow prediction, it also has a certain optimization space: the current mainstream traffic spatial-temporal features extraction schemes focus more on accurately predicting traffic conditions, and lack of interpretable analysis of the impact of different spatial-temporal features on prediction accuracy. Through the analysis of the importance of different spatial-temporal features, the guiding role of spatial-temporal features in traffic prediction tasks is explored; At present, the spatial feature extraction schemes of traffic data tend to use multi-graph convolution, but the definition methods of different schemes for the graph are still not comprehensive. For the schemes with sufficient comprehensive definition, the feature extraction is only for a single road rather than the whole graph. Therefore, defining spatial correlations based on the full-graph scope and utilizing multi-graph convolution for spatial feature extraction remain to be studied. Based on the problems mentioned above, this paper constructs an evaluation module based on the high-dimensional self-attention mechanism to measure the contribution of spatial-temporal features combinations to traffic prediction. And the features extraction module based on multi-graph graph convolution are constructed to capture the spatial-temporal characteristics in traffic data.

The contributions of this paper are summarized as follows:

1. This paper proposes a new spatial-temporal feature extraction and evaluation model based on multi-graph graph convolution and high-dimensional self-attention mechanism, realizing accurate prediction and effect interpretation of traffic data. The model is used to capture and integrate the spatial-temporal correlation characteristics in traffic data, and analyze the impact of different spatial-temporal features on accuracy and visualize the data.

2. In order to clearly distinguish different spatiotemporal characteristics, a variety of spatiotemporal combinations are defined. The temporal features are divided based on the different temporal resolutions, and the spatial features are gathered based on different spatial graph.

3. The improved high-dimensional self-attention mechanism is used to fuse the features of different spatial-temporal combinations, and the influence of different spatial-temporal combinations on the prediction effect is analyzed through the high-dimensional attention score matrix.

4. When extracting spatial-temporal features, based on the traffic data input of each resolution, this paper extracts spatial and temporal features in turn. The spatial features are extracted based on the user-defined multi-dimensional multi graph convolution, and the temporal features are captured through the conventional attention mechanism.

## II. Related Work

In this chapter, some deep learning methods applied to traffic forecasting are discussed. The research on traffic prediction mainly faces the problems of complex traffic dynamic characteristics and nonlinear spatial-temporal correlation. Many scholars provide new ideas to solve the traffic prediction problems by using deep learning technology. Aiming at extracting the spatiotemporal correlation characteristics from traffic data, the deep learning method is mainly divided into single model-based method and multi model fusion method according to the model complexity.

First, the relatively simple single model methods are introduced, and the single model method often pays more attention to capturing the relevant features of a certain dimension of space or time. The following introduce the capture of spatial feature and temporal feature respectively. For spatial correlation features, early scholars focused more on convolutional neural network (CNN), because of the advantages of local perception, weight sharing and the advantages in capturing European spatial correlation features. Toncharoen et al. extracts the spatial features of the data of nodes along the highway through convolutional neural network (CNN) [21], Yao et al. used CNN model to capture the spatial features of the traffic data distributed in the form of region [32], Wang et al. explored 3DResNet and sparse UNet methods to model the spatial correlation of traffic data, among which 3DResNet is a model based on 3D convolutional neural network [33]. However, the traffic data are more irregularly distributed. Compared with CNN, graph convolution neural network (GCN) focusing on non-Euclidean spatial feature extraction has gradually become the mainstream. After the continuous derivation and iteration of GCN, kipf et al. use the first-order approximation of spectral convolution to encode the local graph network, and the GCN learns the non-Euclidean spatial correlation within a certain receptive field through multi-layer stacking. And the computational efficiency and feature extraction performance are balanced [34]. Subsequent applications of GCN are based on this model. In terms of the non-European spatial definition of the traffic network, scholars define the road sensor network or regional network as a graph, that is, the road or region is regarded as a node, and the connection relationship between the roads or regions is regarded as an edge. The non-European spatial correlation characteristics between different nodes are captured through the GCN network. Li et al. propose the DCRNN model to capture the spatial dependence through two-way random walk on the graph through diffusion convolution operation, and use an encoder-decoder architecture with predetermined sampling to capture time dependence and realize multi-step traffic prediction [35]. Chen et al. propose the GCN network to complete the traffic prediction task, and conduct a regression analysis between hyperparameters and GCN performance, so as to balance the stability of prediction accuracy and calculation cost [36]. Jepsen et al. improved the GCN so that the integration of three road network attributes is completed at the same time, that is, the attributes between intersections and good sections. The model performs well in road speed estimation and speed limit classification [37]; Tang et al. and JA et al. introduce an adaptive adjacency matrix mechanism in GCN to dynamically adjust the adjacency matrix based on spatial correlation, so as to improve the real-time performance of capturing spatial features [28, 38].

In terms of nonlinear time feature acquisition of single model, early research was based on RNN and its variants, namely LSTM or GRU, to extract the time features of historical traffic states. Xiao et al. use the bi-directional LSTM model, namely Bi-LSTM, to extract the periodic characteristics in the daily and weekly traffic data, and use the bi-directional characteristics of LSTM to capture the forward and backward traffic flow change trends [39]; Tian et al. used LSTM model to effectively capture the complex time-dependent characteristics of traffic flow in the short-term prediction task of traffic flow [40]. However, RNN networks have the problems of information loss and low computational efficiency when facing the problem of time-dependent feature extraction of long-time series. With the development of attention mechanism in NLP field, attention mechanism shows better feature extraction effect and efficiency for the task of extracting hidden information in the sequence, and shines brightly in traffic prediction tasks. In the task of traffic prediction, different roads or regions can be regarded as a sequence, and the historical traffic conditions of each road or region can also be regarded as a sequence. Therefore, in the process of spatial-temporal feature extraction of traffic data by simply using the attention mechanism, the attention mechanism can often play the role of extracting temporal and spatial features at the same time. Zheng et al. introduce spatial attention mechanism and temporal attention mechanism to extract spatiotemporal features in parallel, and fuse spatiotemporal features through gating fusion module [23]; Xu et al. introduce the spatial transformer and time transformer modules based on attention mechanism. The spatial transformer dynamically modeled the directional spatial dependence with self-attention mechanism to capture the real-time traffic conditions and the directionality of traffic flow. The time Transformer model remote bidirectional time dependencies that span multiple time steps. Finally, two types of Transformer are combined in series into a block to jointly model spatial-temporal correlation to achieve accurate traffic prediction [30].

With the development of deep learning models in the field of traffic prediction, according to the advantages of different models in traffic spatial-temporal feature extraction, the method of integrating different deep learning models to predict traffic status has become a development trend. In short, the model that is more suitable for spatial feature extraction in traffic data and the model that is more suitable for practical feature extraction are combined to achieve better traffic prediction results. The fusion schemes include CNN and RNN fusion, GCN and RNN fusion, GCN and CNN fusion, and CNN, GCN and attention mechanism fusion.

For the scheme of integrating CNN and RNN, Niu et al. propose a new neural network L-CNN in combination with CNN and LSTM. The model can predict the most likely potential passengers of taxi drivers, and L-CNN can be easily extended to the task of road traffic and flow prediction [4]; Cao et al. extract the features of the target road and the surrounding roads with strong correlation through CNN, and completed the periodic acquisition of low-frequency and high-frequency subsequences through LSTM [41]; Bao et al. use three-dimensional convolution and residual elements to construct dynamic spatial feature extraction components to obtain dynamic spatial features. Based on the idea of long-term and short-term memory (LSTM), a new structural element is proposed to extract dynamic temporal features. Finally, the spatial and temporal features were fused to obtain the final prediction results [42]; Zhang et al. propose a new traffic prediction model combining residual network, GCN and LSTM. GCN is used to extract network topology information and LSTM is used to extract time correlation [43].

For the fusion scheme of GCN and RNN, Chen et al. introduce residual recursive network and jump scheme based on GCN model, cooperated with RNN to capture the spatial dependence and time dynamics of traffic, and perform well in traffic prediction [44]; Zhao et al. propose a model combining GCN and GRU, i.e., T-GCN, in order to capture the time-space dependence of the traffic network and predict the traffic at the same time. Among them, GCN is used to learn complex topology to obtain spatial correlation, and GRU is used to learn dynamic changes of traffic data to obtain temporal correlation [45]. Li et al. extracts spatial features through GCN, and the GCN model defines the initial adjacency matrix based on Spearman rank correlation coefficient, and fine tune it during training. The author extracts temporal features through GRU and finally completed the fusion of multi-stream features [46]. Lv et al. encodes the non-European correlation and heterogeneous semantic correlation between roads into multiple graphs, uses GCN to model these spatial correlations, and then uses GRU to learn the dynamic mode of traffic flow, so as to obtain the temporal correlation [22, 47].

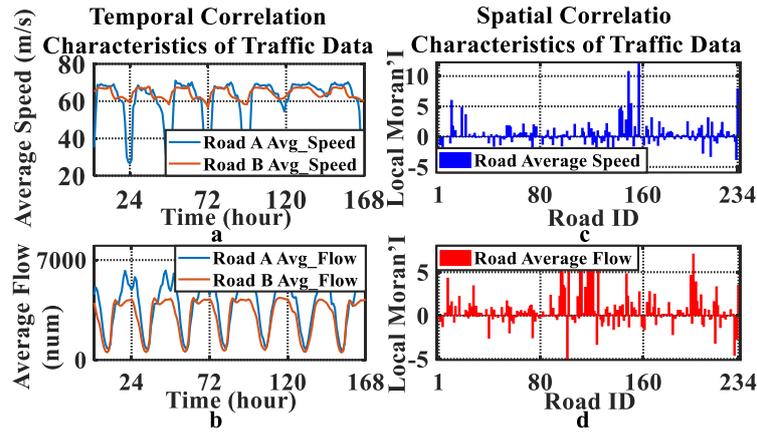

Fig. 1. Spatiotemporal correlation characteristics of the traffic data.

For the fusion scheme of GCN and CNN, Zhang et al. combined GCN and 3D convolution network to complete the short-term passenger flow forecast task. Through multi-graph GCN, three inflow and outflow modes are processed respectively, so as to capture the spatial-temporal correlation of the whole network, and then 3D-CNN is applied to deeply integrate the inflow and outflow information [48]; Zhao et al. introduce GCN based on adaptive adjacency matrix and combine it with CNN to capture spatiotemporal correlation features between nodes [31].

For the combination scheme of CNN, GCN and attention mechanism, Guo et al. propose a new attention based spatiotemporal graph convolution network model to solve the problem of traffic flow prediction. It effectively captures the dynamic spatiotemporal correlation in traffic data through spatiotemporal attention mechanism, and establishes spatiotemporal convolution to capture spatial patterns and temporal characteristics by GCN and CNN separately [49]. Yao et al. propose a new spatiotemporal dynamic network, introduce CNN to learn the dynamic similarity between regions, and design a periodic attention shifting mechanism to capture the long-term periodicity [32]; Bai et al. learn the spatial correlation based on road topology through GCN, and capture the short-term trend by GRU. The attention mechanism to adjust the importance of different time points is introduced, it improves the prediction accuracy by integrating the global time [50]; Zhou et al.

designed a traffic prediction model based on attention mechanism and GCN to extract the spatiotemporal dependence of traffic, in which GCN extracts the sequence correlation and constructs the graph structure based on Spearman rank correlation [51].

## III. PRELIMINARY

### 3.1 Definition 1 (Spatial-Temporal Correlation)

The temporal and spatial correlation of traffic data are focused in this paper. As for the temporal characters of the roads, in Fig.1., the average speed of the floating vehicles and flow of the road in Road A and B located in California are shown in the left-hand chart. In working days, the flow and average speed have obvious regularly peak and trough in the working hour and night duration. Besides, the peak and trough are alleviated on weekends, resulting in the smooth traffic at that time.

Then, in order to measure the correlation of statistical feature of traffic data in terms of spatial distribution, the Global Moran's Index $I_{global}$ and Local Moran's Index $I_{local}$ statistical measures are used for estimating the spatial correlation of traffic data[52, 53]. The Global Moran's Index is used for analyzing the existence of global spatial autocorrelation, and the Local Moran's Index is used for detecting the distribution of spatial autocorrelation of different targets. Specifically, as the distance between targets becomes shorter, the statistical features become more similar, indicating the greater positive Global and Local Moran's Index, conversely, indicating the smaller negative Global and Local Moran's Index. The evaluation is shown in (1-3).

$$I_{global} = \frac{n}{S_0} \times \frac{\sum_{i=1}^{n}\sum_{j=1}^{n} \omega_{ij}(x_i^t - \overline{x}^t)(x_j^t - \overline{x}^t)}{\sum_{i=1}^{n}(x_i^t - \overline{x}^t)^2}, \tag{1}$$

$$I_{local} = (x_i^t - \overline{x}^t) \times \frac{\sum_{j \neq i}^{n} \omega_{ij}(x_j^t - \overline{x}^t)}{\left(\frac{1}{n}\sum_{i=1}^{n}(x_i^t - \overline{x}^t)\right)^2}, \tag{2}$$

$$\omega_{ij} = \begin{cases} 1, & \text{if road } i \text{ is connected to road } j \\ 0, & \text{if road } i \text{ is not connected to road } j \end{cases}, \tag{3}$$

where $I_{global}$ represents the Global Moran's Index, $I_{local}$ represents the Local Moran's Index, $x_i^t$ represents the traffic state of road $i$ at time $t$. In this paper, the traffic state is indicated by the average speed and average flow of the road. $\overline{x}^t$ represents the average value of traffic state of all roads, and $\omega_{ij}$ represents the spatial weight between road $i$ and $j$, and the equation of $\omega_{ij}$ is shown as Eq.3, $n$ represents the number of roads and $S_0$ represents the number of connections between road $i$ and $j$.

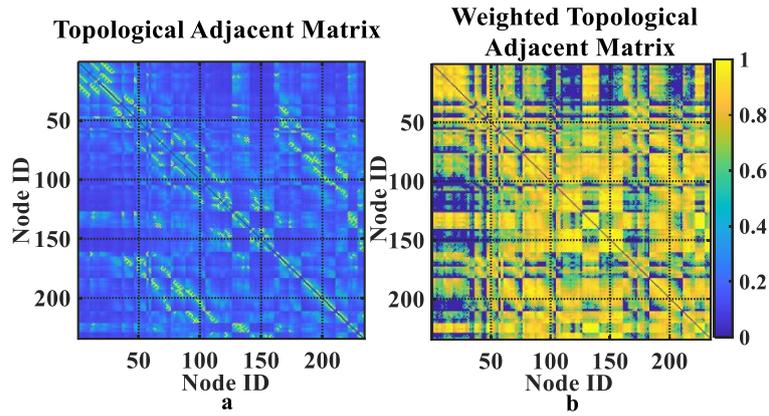

Fig. 2. Heatmap of the topological adjacent matrix and weighted topological adjacent matrix.

As for 234 roads in California, the value of Global Moran's Index is 0.1557 and 0.2461 for road average speed and flow, respectively. The positive value of Global Moran's Index indicates the global spatial positive correlation within the scope of all roads in different traffic state. The Fig.1.b shows the value of Local Moran's Index of traffic state of 234 roads in California. As shown in the chart, most roads show a positive Local Moran's Index in terms of traffic average speed and flow, indicating the distribution of spatial autocorrelation characteristics between different roads in different traffic state.

*3.2 Definition 2 (Road Graph)*

Graph convolution network extracts the spatial characteristics by defining the road network graph. Benefited from different kinds of the road network graph, the non-European spatial features of the road network from different angles are extracted by the graph convolution network. Consequently, the method of definition of the road network graph is absolutely critical. With the accurate coding in aspect of the spatial correlation among all roads in the road network graph, the graph convolution network has stronger learning ability and more accurate prediction effect.

The road network graph is represented as an undirected graph $G = (V, E, W)$, where each node $v_i$ ( $v_{ij} \in V$ ) represents the road in the road network, the number of all roads is represented as $|V| = N$, each edge $e_{ij}$ ( $e_{ij} \in E$ ) represents the correlation between road $v_i$ and $v_j$. The edge weight $w_{ij}$ ( $w_{ij} \in W$ ) of edge $e_{ij}$ represents correlation coefficient between road $v_i$ and $v_j$. The larger the edge weights are, the stronger the correlation

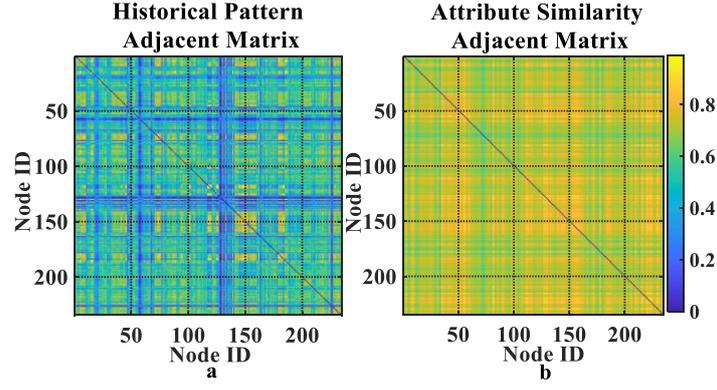

Fig. 3. Heatmap of the historical pattern adjacent matrix and attribute similarity adjacent matrix.

among roads become in a certain aspect. The four types of road network graph are constructed in this paper. Namely, the road topological graph

$G_r$, the road weighted topological graph $G_w$, the traffic historical pattern graph $G_p$ and the road attribute similarity graph $G_s$.

*1) Road Topological Graph:* The road topological graph is expressed as an undirected graph $G_r = (V, E, W_r)$, where the weight $\omega_r(i, j)$ of the edge $e_{ij}$ is the reciprocal of the number of edges traversed by the shortest path from road $v_i$ to road $v_j$ in the road network. Therefore, with the distance among the roads becoming shorter, the corresponding weight in the topological adjacency matrix becomes larger. Then, the adjacency matrix $W_r$ of road topological graph $G_r$ is *expressed as (4), and the visualization of the matrix is shown in the Fig. 2.a.*

$$W_r = \begin{bmatrix} 0 & \omega_r(1,2) & \cdots & \omega_r(1,N) \\ \omega_r(2,1) & 0 & \cdots & \omega_r(2,N) \\ \vdots & \vdots & \ddots & \vdots \\ \omega_r(N,1) & \omega_r(N,2) & \cdots & 0 \end{bmatrix} \quad (4)$$

*2) Weighted Topological Graph:* The road topological graph defines the topological relationship of roads merely, however, the impact on spatial correlation of the length of roads is important as well. Like the road topological graph, the road weighted topological graph $G_w = (V, E, W_w)$ is defined in this paper. The weight $\omega_w(i, j)$ of edge $e_{ij}$ is expressed as (5), where $length(v_m)$ represents the length of all roads in the shortest path from the road $v_i$ to road $v_j$ in the road network. Then, the adjacency matrix $W_w$ of weighted topological graph $G_w$ is expressed as (6), and the visualization of the matrix is shown in the Fig. 2.b.

$$\omega_w(i, j) = \frac{length(v_i) + length(v_j)}{length(v_m)} \quad (5)$$

$$W_w = \begin{bmatrix} 0 & \omega_w(1,2) & \cdots & \omega_w(1,N) \\ \omega_w(2,1) & 0 & \cdots & \omega_w(2,N) \\ \vdots & \vdots & \ddots & \vdots \\ \omega_w(N,1) & \omega_w(N,2) & \cdots & 0 \end{bmatrix} \tag{6}$$

*3) Historical Traffic Pattern Graph:* The road topological graph and weighted topological graph measure the spatial correlation among roads in terms of physical distance, however, the historical traffic pattern graph measures the spatial correlation in the dimension of historical traffic state similarity. The historical traffic pattern graph is represented as $G_p = (V, E, W_p)$. The weight $\omega_p(i,j)$ of edge $e_{ij}$ is the similarity between the historical traffic patterns of road $v_i$ and road $v_j$, and the specific calculation method is as follows:

Firstly, based on the hourly resolution historical traffic condition of road $v_i$, the historical hourly traffic condition sequence $pv_t^i = \{v_{t-\Delta t\_h+1}^i, v_{t-\Delta t\_h+2}^i, \ldots, v_t^i\} \in \mathbb{R}^{\Delta t\_h}$ at time t is formulated, where $\Delta t\_h$ represents the length of the historical traffic mode before the time $t$. If the length of historical traffic mode at daily and weekly resolution are $\Delta t\_d$ and $\Delta t\_w$, the historical daily and weekly traffic condition sequence are $pv_t^i = \{v_{t-\Delta t\_d+1}^i, v_{t-\Delta t\_d+2}^i, \ldots, v_t^i\} \in \mathbb{R}^{\Delta t\_d}$ and $pv_t^i = \{v_{t-\Delta t\_w+1}^i, v_{t-\Delta t\_w+2}^i, \ldots, v_t^i\} \in \mathbb{R}^{\Delta t\_w}$ respectively. Secondly, after counting the historical traffic sequences $pv_t^i$ and $pv_t^j$ of two roads $v_i$ and $v_j$, the DTW (Dynamic Time Warping) is used for calculating the distance between the two historical sequences, expressed as $dist_{p,t}(i,j)$. Thirdly, the weight of historical traffic pattern graph $\omega_p(i,j)$ is calculated by (7) based on the distance $dist_{p,t}(i,j)$. Significantly, the value $\alpha$ in (7) is used for controlling the degree of gain of distance similarity. When the $\alpha$ becomes smaller, the similarity index of the distance becomes larger correspondingly. And the approximate range of value of $\alpha$ is determined by strength of correlation of historical traffic condition. Finally, the weight sequence $\omega_p(i,j)$ among the corresponding roads per hour within a month is obtained, and the adjacency matrix $W_p$ of historical traffic pattern graph $G_p$ is expressed as (8), The visualization of the matrix is shown in the Fig. 3.a.

$$\omega_{p,t}(i,j) = e^{-\alpha \times dist_{p,t}(i,j)} \tag{7}$$

$$W_p = \begin{bmatrix} 0 & \omega_p(1,2) & \cdots & \omega_p(1,N) \\ \omega_p(2,1) & 0 & \cdots & \omega_p(2,N) \\ \vdots & \vdots & \ddots & \vdots \\ \omega_p(N,1) & \omega_p(N,2) & \cdots & 0 \end{bmatrix} \tag{8}$$

*4) Inherent Attribute Similarity Graph:* The graphs defined above extract the spatial correlation in the aspect of topological structure, physical distance and historical time dimension, and the statistical and inherent attributes of roads are also an important aspect of defining the spatial correlation of roads. The attributes of roads are maximum speed, maximum flow and the length of the roads. Therefore, the road attribute similarity graph is expressed as $G_s = (V, E, W_s)$. The weight $\omega_s(i, j)$ of edge $e_{ij}$ is the attribute similarity between road $v_i$ and $v_j$. The calculation method is shown as follows:

Firstly, the maximum flow and maximum average speed of each road are counted per day in a month, and the statistical characteristics are expressed as $\max(flow(v^i)) \in \mathbb{R}^{30}$ and $\max(speed(v^i)) \in \mathbb{R}^{30}$ separately. Due to the different orders of magnitude of flow and average speed of roads, the traffic data of two dimensions above are normalized to the range of [0, 1] by the method of min-max normalization. Secondly, the road attribute sequence $sv_t^i = [\max(flow(v_t^i)), \max(speed(v_t^i)), length(v_i)] \in \mathbb{R}^3$ is formulated corresponding to the road $v_i$ at time $t$. In which, $\max(flow(v_t^i))$ represents the normalized maximum flow of the road $v_i$ on the corresponding day under the time slice $t$. Similarly, $\max(speed(v_t^i))$ refers to the normalized maximum average speed of the road $v_i$ corresponding to the day under the time slice $t$, and $length(v_i)$ represents the length of the road $v_i$. And the attribute distance $dist_{s,t}(i, j)$ between two roads at time $t$ is calculated by sum of squares of each attribute. Thirdly, the weight $\omega_{s,t}(i, j)$ at time $t$ is calculated by the (9). Finally, the weight sequence $\omega_s(i, j)$ corresponding to the roads per hour within a month is calculated. Therefore, the adjacency matrix $W_s$ of attribute similarity graph $G_s$ is expressed as (10), and the visualization of the matrix is shown in the Fig. 3.b.

$$\omega_{s,t}(i, j) = e^{-dist_{s,t}(i, j)} \tag{9}$$

$$W_p = \begin{bmatrix} 0 & \omega_p(1,2) & \cdots & \omega_p(1,N) \\ \omega_p(2,1) & 0 & \cdots & \omega_p(2,N) \\ \vdots & \vdots & \ddots & \vdots \\ \omega_p(N,1) & \omega_p(N,2) & \cdots & 0 \end{bmatrix} \tag{10}$$

### 3.3 Definition 3 (Road Condition Sample)

In order to explicitly represent different types of time periodicity in traffic condition data, the traffic data samples are divided based on three types of time resolution, namely hourly, daily and weekly resolution, and then the different resolution data is inputted into the traffic grade prediction model in parallel. The following is the data division process of three resolutions:

In the hourly resolution channel, the $\Delta h$ hours before the current time $\tau$ are selected as the input data, and the input is expressed as (11):

$$X_h = (X_{\tau - \Delta h + 1}, X_{\tau - \Delta h + 2}, ..., X_\tau) \tag{11}$$

In the daily resolution channel, in order to predict the traffic conditions of the next $t_p$ time slice from the current time $\tau$, the $\Delta d$ days before $t_d = \tau + t_p - 24$ is intercepted as the input data, and the input is expressed as (12):

$$X_d = (X_{\tau-len\_d+1}, X_{\tau-len\_d+2}, ..., X_{t_d}) \tag{12}$$

In the weekly resolution channel, similarly, the input data is represented by the historical traffic data of $\Delta w$ weeks before time $t_w = \tau + t_p - 24 \times 7$, and then the input is defined as (13):

$$X_d = (X_{\tau-len\_d+1}, X_{\tau-len\_d+2}, ..., X_{t_d}) \tag{13}$$

Finally, $X_h$, $X_d$ and $X_w$ are passed as inputs to spatial-temporal feature extraction layer.

### 3.4 Definition 4 (Traffic Condition Grades)

The predicted traffic conditions are expressed in the form of grades, so as to more intuitively and comprehensively represent the traffic conditions. In this paper, the characteristics of traffic conditions $X$ are inputted into the Self-Organizing Mapping neural network (SOM), classifying the traffic data so as to obtain the traffic grades $Y$ of the corresponding roads, $Y \in \{1, \cdots, Class\}$ represents the value range of traffic grade, $Class$ represents the number of the traffic grade. The classification process is as follows: Firstly, the SOM is trained on all samples of traffic conditions, and based on the trained network, the grades of traffic conditions are divided in test process. The pseudo code of training process and test process of SOM algorithm is shown in Table 1-2.

Table 1 The training process of SOM algorithm

---

**Algorithm 1 Training process of Self-Organizing Mapping neural network (SOM)**

---

**input**: The normalized traffic condition sample $X$; The number of grades of traffic state $Class$; The size of dimension $[n\_row, n\_col]$, and $n\_row \times n\_col = Class$.

**output**: Trained SOM network.

1 Initialize $iter$ with 1; Initial *MaxIter* with 200;

    Initial the size of distribution of the nodes with $n\_row$ rows and $n\_col$ columns;

    Initial the weight vector of nodes $W = \{\omega_{jk}, j = 1, \cdots, Class, k = 1, 2, 3\}$ with the random number in range of [0, 1];

    Initialize the learning rate $f_{learn}(1) = l_0$ with 0.1;

    Initialize the neighborhood radius function $f_{neighbor}(1) = n_0$ with 3.

2 **while** $iter < MaxIter$ **do**:

3 **for** $i \leftarrow 1$ **to** $T \times N$ **do** (for each sample in the traffic condition sample):

4   Choose the traffic condition sample $x_i \in X$.

5 **for** $j \leftarrow 1$ **to** *Class* **do** (for each weight in the hidden layer of neural network):

6    Choose the weight $\omega_j$ of the node $j$.

7    Calculate the Euclidean distance between the traffic condition sample $x_i$ and the weight of

     neural network $\omega_j$:

$$d_{ij}(x) = \sqrt{(x_i - \omega_j)^2}$$

8    Find the closest node and consider it as the winning node.

9   **for** $k \leftarrow 1$ **to** $n_0$ **do**:

10    Update the weight of the winning node and the neighbor nodes less than $n_0$ from the winning node.

$$\omega_k = \omega_k + f_{learn}(iter) \times f_{neighbor}(iter) \times (x_i - \omega_k)$$

11    Update the number of iteration, learning rate and neighborhood radius function:

$$f_{neighbor}(iter+1) = n_0 \times \exp(-iter / t_1), t_1 = MaxIter / \log(n_0) \quad f_{learn}(iter+1) = l_0 \times \exp(-iter / t_2), t_2 = MaxIter$$

$$iter = iter + 1$$

12    **end for**

13   **end for**

14  **end for**

15 **end while**

Table 2 The Testing process of SOM algorithm

**Algorithm 1 Testing process of Self-Organizing Mapping neural network (SOM)**

**input**: The normalized traffic condition sample $X$; Trained SOM network.

**output**: The traffic grades $Y$ corresponding to the traffic condition sample $X$.

1 **for** $i \leftarrow 1$ **to** $T \times N$ **do** (for each sample in the traffic condition sample):

2   Choose the traffic condition sample $x_i \in X$.

3   **for** $j \leftarrow 1$ **to** $(n\_row \times n\_col)$ **do** (for each weight in the hidden layer of neural network):

4    Choose the weight $\omega_j$ of the node $j$.

5    Calculate the Euclidean distance between the traffic condition sample $x_i$ and the weight of

     neural network $\omega_j$:

$$d_{ij}(x) = \sqrt{(x_i - \omega_j)^2}$$

6    Find the closest node $\omega_m, m \in \{1, \cdots, Class\}$ and determine the traffic grade of the sample $x_i$

     as grade $m$.

7   **end for**

8 **end for**

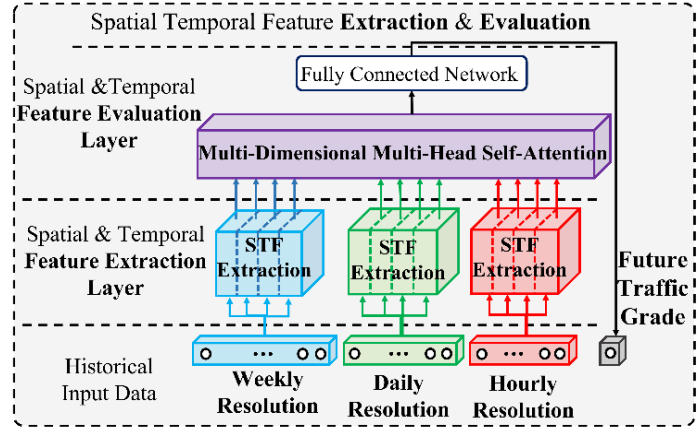

Fig. 4. The architecture of DL-STFEE model

Problem statement: the goal of urban traffic grade prediction based on road network is to learn a prediction function under the observed values of historical traffic conditions, namely road average speed and flow, of all roads $X \in \mathbb{R}^{N \times \Delta t \times 2}$ , and the function can map the historical traffic conditions at different resolutions at the current time $\tau$ to the future traffic condition grade $Y_{\tau + t_p}$ at the moment $\tau + t_p$ on the premise of $G_r$ , $G_w$ , $G_p$ and $G_s$ , the function is shown in (14).

$$Y_{\tau + t_p} = f(G_r, G_w, G_p, G_s; X_h, X_d, X_w)$$ (14)

## IV. Method

In this chapter, the architecture of DL-STFEE model is elaborated, as shown in Fig. 4. The model is mainly composed of four parts: input layer, spatial temporal feature extraction layer, spatial temporal feature fusion layer and fully connection layer. Firstly, the input layer divides the historical traffic condition into three resolutions, namely hourly, daily and weekly, so that the model is more conducive to learn the different periodic laws; Secondly, the spatial temporal feature extraction layer is composed of three parallel multi-graph graph convolution model combined with the conventional attention mechanism. The graph convolution model captures the spatial correlation features in the traffic data. Specifically, the complex spatial correlation features are extracted by graph convolution model from different angles based on the predefined four adjacency relationships, and the correlation of the time dimension is extracted by the conventional attention mechanism;

Thirdly, the spatial and temporal features output from the spatial temporal feature extraction layer are input to the spatial temporal combination fusion layer, and the spatial  temporal commination fusion layer is constructed by a multi-head attention model based on graph. The spatial temporal combinations of temporal features with different temporal resolutions and spatial features under different distributions are fused; Finally, the fused spatial temporal combination features are inputted into the full connection layer, so as to output the future traffic grade of each road.

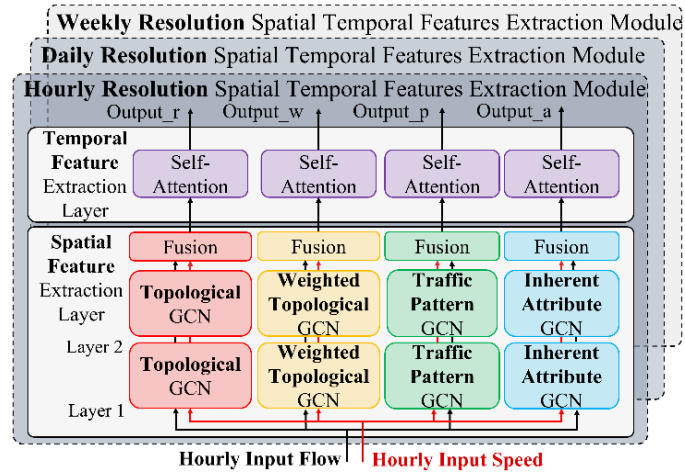

Fig. 5. The structure of spatial-temporal feature extraction layer.

### 4.1 Spatial-Temporal Feature Extraction Layer

The spatiotemporal feature extraction layer is composed of a spatial feature extraction layer and a temporal feature extraction layer. The specific structure is shown in Fig. 5. First, the multi-channel traffic data are inputted into the spatial feature extraction layer to extract different aspects of spatial features. The spatial feature extraction layer is composed of two layers of multi-graph convolution and fusion operations to complete the extraction of the full-map road spatial features and fuse with multi-channel features; secondly, the spatial features extracted at each angle are inputted to the temporal feature extraction layer. The extraction layer consists of a conventional self-attention mechanism, and it can effectively capture the temporal correlation between different historical data. Finally, the spatiotemporal features under the channels based on different graph types are used as the output of the spatiotemporal feature extraction layer.

*1) Graph Convolution for Traffic:* The traffic network usually presents the structural form of graph with non-Euclidean properties. Aiming to not only more in line with the actual road distribution, but also more conductive to model learning the spatial features of the road, the road network is represented as the mathematical method of graph. For the traffic data in graphical form, compared with conventional convolution model, the graph convolution is used for extracting the spatial non-European correlation of traffic data in this paper.

Graph Convolution Network: In order to simplify the graph convolution model, Kip F and Welling proposed the third generation GCN[34], and the characteristics of this generation of GCN is fitting the convolution kernel by the first-order approximation of Chebyshev polynomial. Although each layer of graph convolution only considers the direct neighborhood in graph, the grid depth is reduced by stacking multi-layer local map convolution layers, so as to simplify the model and build a deeper architecture to expand the receptive field. The constraints $K=1$ and $\lambda_{max} \approx 2$ is added to the graph convolution, so that the formula can be simplified to (15):

$$\Theta * \varsigma x = \Theta(L)x \approx \theta_0 x + \theta_1(\frac{2}{\lambda_{max}}L - I_n)x$$
$$\approx \theta_0 x - \theta_1(D^{-1}WD^{-\frac{1}{2}})x \tag{15}$$

where $\theta_0$ and $\theta_1$ are two shared parameters of convolution kernel. In order to further simplify the model and increase numerical stability, $\theta_0$ and $\theta_1$ are replaced by a single parameter $\theta$, namely

$\theta_0 = \theta = -\theta_1$. Therefore, the single convolution kernel parameter per channel has only one parameter $\theta$. $W$ and $D$ are normalized to $\tilde{W} = W + I_N$ and $\tilde{D}_{ii} = \sum_j \tilde{W}_{ij}$, separately. The graph convolution can then be replaced by (16):

$$\Theta * \varsigma x = \theta(I_n + D^{-\frac{1}{2}}WD^{-\frac{1}{2}})x$$
$$= [\theta(\tilde{D}^{-\frac{1}{2}}\tilde{W}\tilde{D}^{-\frac{1}{2}})x] \in \mathbb{R}^{N \times 1} \tag{16}$$

where, among them, the convolution kernel is $\theta \in \mathbb{R}$, the normalized adjacency matrix is $(\tilde{D}^{-\frac{1}{2}}\tilde{W}\tilde{D}^{-\frac{1}{2}}) \in \mathbb{R}^{N \times N}$, and the input signal is $x \in \mathbb{R}^{N \times 1}$.

Considering that the characteristics of input data are generally more than one dimension, in order to extract multidimensional input data, the graph convolution operation based on one-dimensional input $x \in \mathbb{R}^{N \times 1}$ can be generalized to multi-dimensional input $X \in \mathbb{R}^{N \times C}$. The operation of the graph convolution is shown in (17):

$$Z = (\tilde{D}^{-\frac{1}{2}}\tilde{W}\tilde{D}^{-\frac{1}{2}})X\Theta \tag{17}$$

where the graph convolution kernel is $\Theta \in \mathbb{R}^{C \times F}$, the normalized adjacency matrix is $(\tilde{D}^{-\frac{1}{2}}\tilde{W}\tilde{D}^{-\frac{1}{2}}) \in \mathbb{R}^{n \times n}$, and the input signal is $X \in \mathbb{R}^{N \times T}$. $T$ represents the dimension of features of the input signal, and $F$ represents the number of convolution kernels, namely the characteristic dimension of the output signal. Specifically, for traffic condition forecasting task, the input data consists of the road graph and traffic conditions under $T$ timestamps. Assuming the input data is $X \in \mathbb{R}^{N \times T}$ per timestamp, namely the $T$-dimension feature matrix of $N$ road nodes, and the input of graph data is $\tilde{W}$, namely the adjacency matrix introduced in chapter III. 3.2. Therefore, the graph convolution based on multi convolution kernel of multi-dimensional input is calculated by applying $F$ shared convolution kernel $\Theta$ and normalized adjacency matrix $(\tilde{D}^{-\frac{1}{2}} \tilde{W} \tilde{D}^{-\frac{1}{2}}) \in \mathbb{R}^{N \times N}$ to input $X \in \mathbb{R}^{N \times T}$ in parallel.

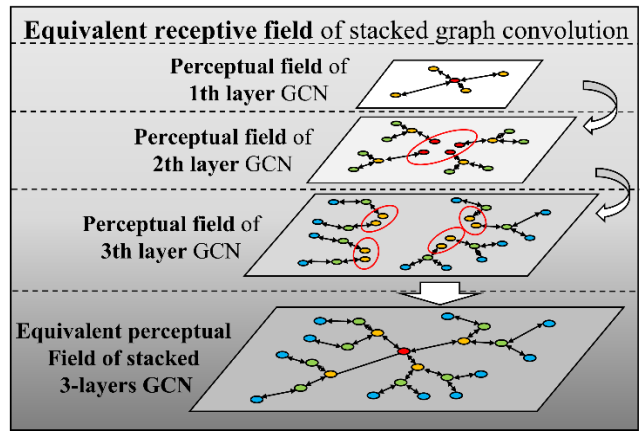

Fig. 6. Equivalent receptive field of stacked graph convolution.

Next, the advantage of first-order approximate graph convolution network is elaborated. The receptive field of the traffic features extracted by graph convolution model is shown in the Fig. 6. The nodes of graph in figure represent the roads, and the edges of graph represent the connectivity among roads. The receptive field of the single-layer first-order approximate graph convolution can only cover the first-order neighbors of the central road, but the third-order receptive field is can be achieved by stacking three layers' graph convolution of first-order approximation. Aiming to cover the same receptive field, the single layer graph convolution of third-order approximation is also a method. However, the former is better in number of parameters and computational efficiency. In this paper, the first-order approximate graph convolution model with two layers is designed to capture the spatial correlation of each roads' second-order neighbors.

3) Multi-Dimensional and Multi-graph Graph Convolution: Considering the diversity of the input traffic condition data, the traffic prediction task based on one kind of data alone is not comprehensive. In order to predict the traffic condition by making full use of the existing data, and consider the impact of various traffic data on the prediction accuracy, the variety of feature embedding are extracted in this paper. Moreover, considering the certain correlation among the different types of traffic data, the public information of heterogeneous traffic data is extracted by shared multi-channel parameter graph convolution model.

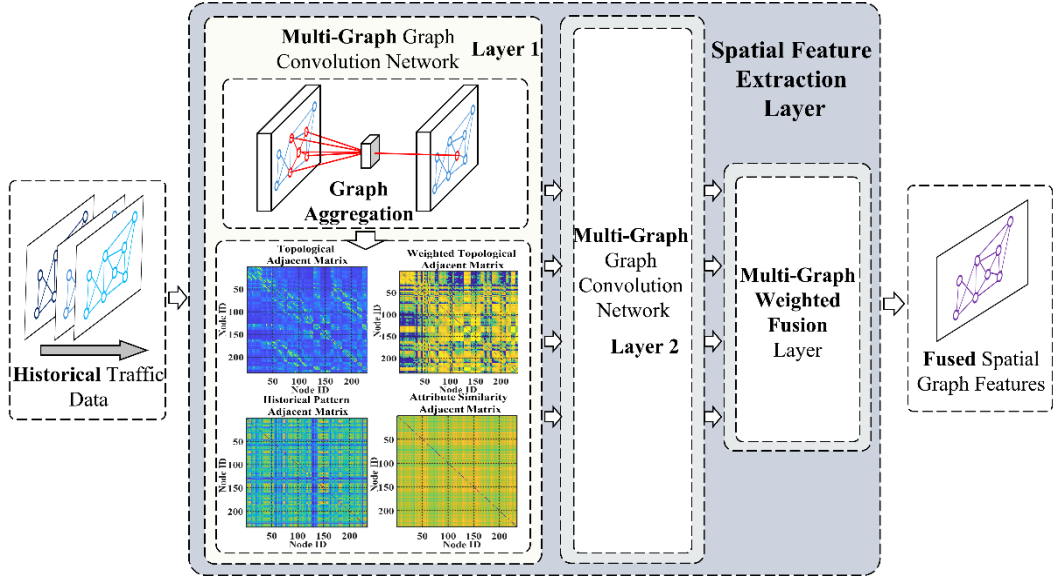

Fig. 7. Multi-channel multi-graph convolution attention module structure diagram

The multi-channel and multi-graph graph convolution proposed in this paper is explained from two parts, namely the angle of multi-channel and multi-graph. For the multi-channel's point of view, the double-layer graph convolution with shared multi-channel parameter is designed, so that the multi-channel input data can be operated by graph convolution model, and the model also extracts the shared features from the different type of input data. In this paper, the multi-channel inputs are average speed and average flow of roads separately. For the angle of multi-graph of graph convolution, namely the four graphs defined in chapter III. 3.2, the graph convolution is operated based on every graph data, so that the spatial correlation features of traffic data are extracted in different angles, the double-layer graph convolution is used in this paper.

As shown in Fig. 7, based on every kind of graph, the graph convolution of input data from different channels is calculated in parallel, then the feature of fusion is calculated by the fusion layer, so that the merged feature embedding is obtained as the output of the graph convolution of the double-layer and two-channel under a certain graph. The above operation is used in parallel in the multi-channel traffic data based on every type of graph data, finally the output in different type of graphs is obtained. Next, the topological graph-based graph convolution computation process based on hourly resolution traffic data is shown below.

First, based on topological graph $G_r = (V, E, W_r)$, the process of calculation of shared-GCN to extract the feature of the traffic data is shown in (18).

$$Z_{r\_speed\_\Delta h}^{(l)} = \mathrm{Re}\,LU(\tilde{D}_r^{-\frac{1}{2}}\tilde{W}_r\tilde{D}_r^{-\frac{1}{2}}Z_{r\_speed\_\Delta h}^{(l-1)}W_r^{(l)}) \tag{18}$$

where, $W_r^{(l)}$ is the weight matrix of $l$ layer of the shared GCN, and $Z_{r\_speed}^{(l-1)}$ is output embedding of average road speed of layer $l-1$, $Z_{r\_speed}^{(0)} = X_{speed\_\Delta h}$, that is, the input of road speed feature of all roads at hourly resolution. For another feature road flow, the same weight matrix $W_r^{(l)}$ at each layer of the shared GCN is used for extracting the shared feature, as shown in (19) below.

$$Z_{r\_flow\_\Delta h}^{(l)} = \mathrm{Re}\, LU(\tilde{D}_r^{-\frac{1}{2}} \tilde{W}_r \tilde{D}_r^{-\frac{1}{2}} Z_{r\_flow\_\Delta h}^{(l-1)} W_r^{(l)}) \tag{19}$$

where $Z_{flow\_\Delta h}^{(l)}$ is the output embedding of road flow feature of layer $l$, and $Z_{r\_flow}^{(0)} = X_{flow\_\Delta h}$, that is the input of road flow feature of all roads at hourly resolution. The shared weight matrix can filter the shared features from the two feature spaces. According to the input data in different channel, we can get two output embeddings $Z_{r\_SPEED\_\Delta h}$ and $Z_{r\_FLOW\_\Delta h}$.

After the convolution operation of the second layer graph convolution, the trainable weight vectors $W_{r\_speed\_\Delta h}$ and $W_{r\_flow\_\Delta h}$ are formulated and Element-wise multiplied with the feature embedding of the output of the graph convolution to obtain the combined output, as shown in (20).

$$Z_{r\_\Delta h} = W_{r\_speed\_\Delta h} \circ Z_{r\_SPEED\_\Delta h} + W_{r\_flow\_\Delta h} \circ Z_{r\_FLOW\_\Delta h} \tag{20}$$

For the angles of multi-graph, according to chapter III. 3.2, the four kinds of graphs are defined, namely, encoding the relationship between roads from the aspects of road topology relationship, traffic mode relationship and road attribute. The road topological graph at hourly resolution is assumed below, according to III. B. the corresponding graph convolution output embedding $Z_{r\_\Delta h}$ is obtained. similarly, the output embedding $Z_{w\_\Delta h}$, $Z_{p\_\Delta h}$ and $Z_{s\_\Delta h}$ under the convolution of multi-channel graph corresponding to weighted topology graph, historical pattern graph and attribute similarity graph are also obtained. Similarly, based on daily resolution and weekly resolution, the output embedding based on the graph convolution with multi-channel input and multi-graph data are also calculated.

4) General Self-attention Module: The spatial features of traffic data are extracted by multi graph convolution, and the feature vectors of different regions are obtained. In order to further capture the time correlation of different regions, this paper adopts the ordinary attention module to obtain the feature vector of time dimension. The feature data obtained from the above multi graph convolution module is $x_{ij} \in \mathbb{R}^{n \times d}, i = 1, 2, 3, j = 1, ..., 4$, where $i$ represents three different resolutions, $j$ represents the graph convolution under the definition of four different graphs, $n$ represents the number of roads, and $d$ represents the feature number of each road, so as to obtain 12 different scheme combinations. For each combination, the self-attention module is used for time feature extraction.

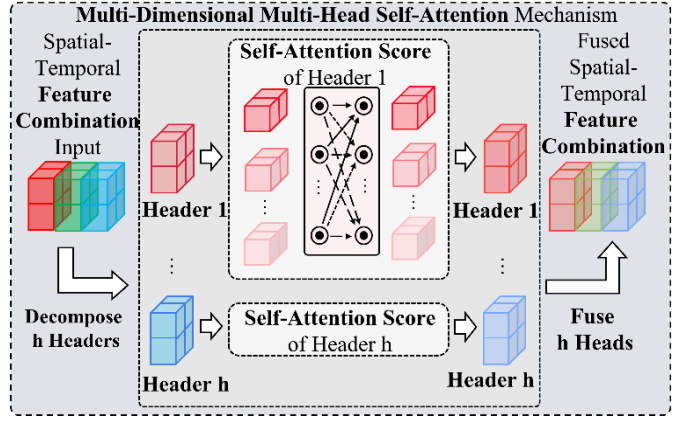

Fig. 8. The Process of Calculation of the multi-head high-dimensional self-attention module

$x'_{ij} \in \mathbb{R}^{d \times n}$ is obtained by transposing $x_{ij}$, in which the time feature dimension represents the number of elements and the area feature dimension represents the characteristics of elements. Then, the operation of (21) is performed,

$$Attention(Q, K, V) = soft \max(\frac{QK^T}{\sqrt{d_k}})V \qquad (21)$$

where $Attention(.)$ represents the conventional attention operation. By the operation of scaled dot product and softmax to Q and K, the weight matrix $W = soft \max(\frac{QK^T}{\sqrt{d_k}})$ is obtained. And by multiplying the weight matrix $W$ by $V$, the operation of conventional attention is achieved. $Q = K = V = x'_{ij}$ represents the time feature vector sequence, $x'_{ij} \in \mathbb{R}^{d \times N}$ is obtained by transposing $x_{ij}$. $d_k$ represents the feature dimensions of the $Q$ and $K$. For avoiding the result of the dot product too large and improving computing efficiency, the scaled operation is used in dot product, and the adjustment factor is $\sqrt{d_k}$. Finally, the results are transposed to obtain different time features under each road, as shown in (22-23).

$$\hat{x}_{ij} = Attention(x'_{ij}, x'_{ij}, x'_{ij}) = soft \max(\frac{x'_{ij}{x'_{ij}}^T}{\sqrt{d_k}})x'_{ij} \qquad (22)$$

$$x''_{ij} = \hat{x}'_{ij} \qquad (23)$$

In this way, the traffic data is decoupled and reorganized from three perspectives: the resolution of the input data, the dimension of the input data and the formulation of the graph, thus the combinations of different angles of features in the traffic data are obtained. In detail, the three types of time resolution combinations (hourly, daily and weekly) and four combinations of graph (topology and weighted topology, historical mode and road attribute) are constructed, with a total of twelve combinations. Next, the impact of these combinations on the prediction accuracy needs to be determined by the spatial temporal combination layer.

*4.2 Spatial-Temporal Combination Layer*

The spatiotemporal combination fusion layer is composed of high-dimensional self-attention mechanism and full connection layer, and its structure is shown in the Fig. 4. Firstly, different combinations of spatiotemporal features are obtained through the spatiotemporal feature extraction layer, and the combinations are input into the high-dimensional self-attention mechanism. Different combinations of features are fused in the high-dimensional self-attention mechanism, and the weights between different combinations are obtained according to the attention score matrix. Specifically, the evaluation of different spatiotemporal feature combinations in this paper is mainly based on the attention score matrix. Finally, the spatiotemporal features fused by the high-dimensional self-attention mechanism are input into the full connection layer, so as to complete the output of the prediction grades of all roads.

*1) High Dimensional Self-Attention Mechanism:* Through the multi-channel and multi graph convolution and attention mechanism, the features of the data are extracted from three angles, so as to obtain the feature embedding of twelve combinations. We parallel the outputs under different combinations to obtain the feature representation of different combinations $X \in \mathbb{R}^{T_p \times N \times d}$ (where $T_p$ represents the sequence length, in this paper $T_p = 12$, $n$ represents the number of roads, $d$ represents the number of features of each road, and the values are different according to different resolutions).

In order to further extract the interaction information of different combinations and measure the impact of different combinations on the prediction accuracy, we fuse the spatial temporal combinations through the multi head high-dimensional self-attention layer. The specific process is shown in Fig. 8. Considering that the input feature embedding is the high-dimensional data output through graph convolution, compared with the normal input data of the autonomous force layer in the form of vector, the input data of the high-dimensional self-attention layer is in the form of matrix (equivalent to adding a dimension). Therefore, this paper introduces the multi head high-dimensional autonomous force module, which is similar to the traditional autonomous force mechanism. We regard the first dimension as the sequence length and the second and third dimensions as the characteristic matrix of an element in the sequence.

Firstly, the matrix ( $Q \in \mathbb{R}^{T_p \times N \times d}$ , $K \in \mathbb{R}^{T_p \times N \times d}$ and $V \in \mathbb{R}^{T_p \times N \times d}$ ) of query, key and value are defined, and the matrices above are collectively referred to as the functional matrices. By performing linear transformation on the second dimension, i.e., the number dimension of features of the roads, of traffic embedded data $X \in \mathbb{R}^{T_p \times N \times d}$ and decomposing it into multiple headers, the functional matrices are obtained, as shown in (24-26).

$$Q_i = [reshape(W_Q X_{fe})]_i \tag{24}$$

$$K_i = [reshape(W_K X_{fe})]_i \tag{25}$$

$$V_i = [reshape(W_V X_{fe})]_i \tag{26}$$

where $W_Q \in \mathbb{R}^{N \times N}$ , $W_K \in \mathbb{R}^{N \times N}$ and $W_V \in \mathbb{R}^{N \times N}$ represent the learnable parameter matrix, $h$ represents the number of headers, $reshape(.)$ represents the reshape operation of the matrix, i.e., converting the shape of $[T_p \times N \times d]$ to $[T_p \times h \times \frac{N}{h} \times d]$ . $Q_i \in \mathbb{R}^{T_p \times \frac{N}{h} \times d}$ , $K_i \in \mathbb{R}^{T_p \times \frac{N}{h} \times d}$ and $V_i \in \mathbb{R}^{T_p \times \frac{N}{h} \times d}$ represent the three functional matrices of the corresponding header $i$ obtained through the linear mapping and reshape operation separately.

Then, dot product attention operation is used for representing the operation of multi-head high-dimensional self-attention, as shown in (27-29) below.

$$a_i^{t,t'} = soft \max(f(Q_i^t, K_i^{t'})) = \frac{\exp(f(Q_i^{t \times (d_q \times d)}, K_i^{t' \times (d_k \times d)}))}{\sum_{t' \in T_p} \exp(f(Q_i^{t \times (d_q \times d)}, K_i^{t' \times (d_k \times d)}))} \tag{27}$$

$$head_i = Attention(Q_i, K_i, V_i) = \sum_{t \in T_p} a_i V_i^t , \quad i \in [1, \cdots, h] \tag{28}$$

$$X_{fe} = MultiHead(Q, K, V) = Concat(head_1, \cdots, head_h) W^O \tag{29}$$

where $a_i^{t,t'} \in \mathbb{R}^d \in A$ ( $A \in \mathbb{R}^{h \times T_p \times d}$ , $i \in \{1, \cdots, h\}$ ) represents the normalized weight matrix between combination $t$ and combination $t'$ of $i$ -th header, $t \in T_p$ , the second dimension number of $Q$ , $K$ and $V$ are $d_q = d_k = d_v = \frac{N}{h}$ , $head_i$ represents the feature matrix of header $i$ . Finally, the final output of the high-dimensional multi-head self-attention model is obtained by splicing and linear mapping of all head matrices.

In DL-STFEE architecture, by using one layer of multi-head high-dimensional attention operation, the feature information embedding among different combinations, the importance score among different combinations and the importance score of impact on the results are obtained.

*2) Fully Connection Layer:*

Not only to use the features extracted by the backbone network for prediction tasks and complete the matching of output size, but also to increase the depth of the network layer to further strengthen the learning ability of the model, the fully connected layer is used as the head part of the overall network. Specifically, the output tensor of multi-head high-dimensional self-attention $X_{fc} \in \mathbb{R}^{T_p \times N \times d}$ is transposed and stretched into the two-dimensional vector $X'_{fc} \in \mathbb{R}^{N \times (T_p \times d)}$ and fed to the full connection layer. The process of calculation of the full connection layer can be expressed as (30).

$$Y_{fc} = \mathrm{Re}\,LU(W_{fc}X'_{fc} + B_{fc}) \tag{30}$$

where $Y_{fc} \in \mathbb{R}^{N \times Class}$ represents the output of the fully connection layer, $N$ represents the number of the roads, $Class$ represents the number of the traffic congestion grade, $W_{fc}$ and $B_{fc}$ represent the weight matrix and deviation matrix in turn.

*4.3 Loss Function*

In this paper, our DL-STFEE is trained to predict traffic by minimizing the Negative Log Likelihood Loss (NLLLoss) between the estimated level and the ground real level. In the multi-level classification task, the estimated level matrix is obtained by performing the negative logarithm softmax operation on the output matrix, and then the obtained estimated level matrix and the real level are input into the NLLLoss function to calculate the loss value, so as to complete the back propagation. The loss function is calculated as the average value of the loss function of all training data. For $i$ -th element of the output, find the value of the corresponding position in the estimated level vector according to its real category label $k$ , and then perform natural logarithm and negative operation on it. The output value is used as the loss function value of this category. Finally, the loss function values of all elements are averaged as the output of NLLLoss. The calculation process is as shown in (31-32).

$$y_{pred}^{i,j} = soft\max(y_{fc}^{i,j}) = \frac{\exp(y_{fc}^{i,j})}{\sum_j \exp(y_{fc}^{i,j})} \tag{31}$$

$$loss(Y_{pred}, Y_{true}) = -\frac{1}{N}\sum Y_{true} * \log(Y_{pred}) \tag{32}$$

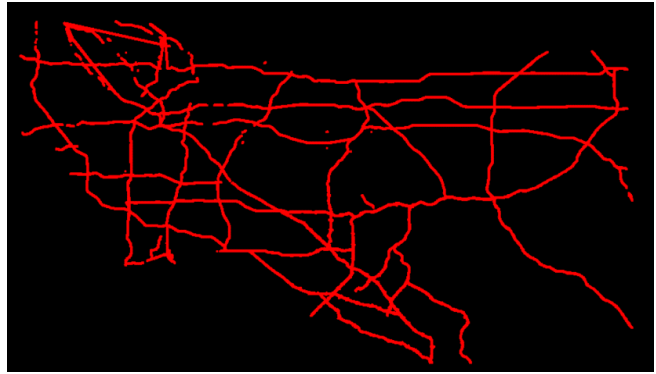

Fig. 9. Traffic data visualization diagram

where $a_{i,j}$ represents the normalized weight of grade $j$, $(j \in Class)$ of road $i$, $(i \in N)$. $N$ represents the number of the road. $Class$ represents the congestion grade of the road, the grade label $k \in Class$. $loss(a_i, label = k)$ represents the loss value of predicting the grade of road $i$ as grade $k$.

## V. Experiment

### 5.1 Experimental Setting

We evaluate our approach based on a real-world data set, i.e., $PEMSD\ 07/08/12$. All experimental evaluations were performed on the window 10 system and NVIDIA RTX 2080. We implement our model and deep learning baseline on the Pytoch platform, and the traditional database based on machine learning is implemented by scikit-learning. Other parameters are shown in Table 3.

*1) Data Description:* $PEMS\ 07/08/12$ was collected by the California Department of transportation. We selected 2547 sensors as data sources in the corresponding three districts of the highway system in California. The period of $PEMS\ 07/08/12$ is from January 1, 2020 to January 31, 2020. Each sensor measures the average road speed and flow every hour.

*2) Data Processing:* The data measurement interval of the data set is 1 hour, so each sensor of the road map contains 744 data points. According to the road network structure, we average all sensor data of the same road as the traffic data of the road. There are 234 roads in total. See Fig. 9. for the specific distribution. We set the resolution set $P \in \{hour, day, week\}$ for *PEMSD* 07/08/12 and map the input data to [0,1] using min max normalization.

*3) Parameter Setting:* In the experiment, the super parameters according to the existing data set are formulated, and the specific parameters are shown in Table 3.

Table 3 Detailed Parameter Setting

| Detailed Parameter Setting | | | |
|---|---|---|---|
| Number of roads | 234 | Number of 1[th]GCN Hidden layer | 32 |
| $W_h$ | 24 | Number of 2[th]GCN Hidden layer | 32 |
| $W_d$ | 7 | Number of fully connected hidden layer | 32 |
| $W_w$ | 3 | Number of road grades | 5 |
| DTW attenuation rate of velocity | 1e-2 | Optimizer | ADAM |
| DTW attenuation rate of flow | 1e-4 | Learning rate | 1e-3 |
| Size of training set | 240 | Batch Size | 16 |
| Size of validation set | 80 | Epoch | 500 |
| Size of test set | 80 | | |

*4) Evaluation Index:* In this paper, the accuracy and weighted kappa coefficient are used as the evaluation indicators of the prediction effect. The accuracy represents the ratio of the number of cases in the prediction grade that are the same as the actual grade to the total number of prediction cases, and represents the accuracy of prediction; The quadratic weighted kappa coefficient indicates the consistency between the grade results of traffic prediction and the real grade results, that is, not only the accuracy of the prediction results, but also the deviation degree of the prediction results. The calculation is based on the confusion matrix, and the value is between -1 and 1. The closer the value is to 1, the higher the consistency of the prediction grade results. The calculation method of accuracy and weighted kappa coefficient are shown as (33-35).

$$Accuracy = \frac{1}{n}\sum_{t=1}^{n}(1, \ if \ v_t = \tilde{v}_t \ else \ 0) \tag{33}$$

$$Quadratic \ Weighted \ Kappa = \frac{P_o - P_e}{1 - P_e} \tag{34}$$

$$\begin{cases} P_o = \sum_{i=1}^{Class}\sum_{j=1}^{Class}\omega_{i,j}p_{i,j} \\ P_e = \sum_{i=1}^{Class}\sum_{j=1}^{Class}\omega_{i,j}p_{i.}p_{.j} \\ \omega_{i,j} = 1 - \left(\frac{i-j}{Class-1}\right)^2 \end{cases} \tag{35}$$

where, the accuracy calculation process is shown in (30), $v_t$ represents the real grade, $\tilde{v}_t$ represents the prediction grade, and $n$ represents the number of predicted values. The Equation (31-32) represent the process of calculation of the weighted kappa coefficient, $p_{i,j}$ represents the quantity distribution matrix that discriminates grade $i$ as grade $j$, that is, $p$ is the confusion matrix, $p_{i,j}$ represents the expected quantity distribution matrix that discriminates grade $i$ as grade $j$ and its calculation method is shown in (32). $\omega_{i,j}$ represents the gain weight, $Class$ represents the number of grades.

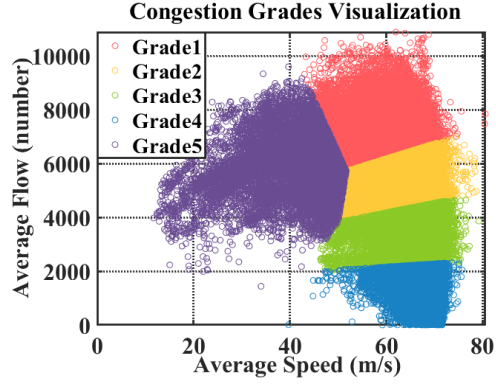

Fig. 10. Visualization of traffic congestion classification results

*5) Congestion Grade:* According to the above, we use SOM method to get the traffic congestion level of the corresponding roads according to the traffic data of all roads at all times. The visualization results are as follows. As can be seen from Fig. 10., the distribution boundaries of different congestion levels are very clear, which also shows that the characteristics of different congestion levels are very obvious and more representative.

*5.2 Experiment 1: Effect of Combination*

According to the chapter IV. 4.2, the different combinations have different perspectives in extracting the spatiotemporal features of traffic data. As shown in the Fig. 11.a, the combination symbols are composed of lowercase letters and subscripts. The lowercase letters represent the extraction of spatial features based on different graph angles, $r$ representing the topology map, $w$ representing the weighted topology map, $p$ represents the traffic pattern map, and $s$ represents the road attribute map. The subscripts indicate that temporal features are extracted based on angles of different temporal resolutions, where $h$ represents the hourly resolutions, $d$ represents the daily resolutions, and $w$ represents the weekly resolutions. Aiming at measuring the importance among different combinations, the graph attention weight $A \in \mathbb{R}^{h \times T_p \times T_p \times d}$ of the spatial temporal feature confusion layer is processed and analyzed. With the greater the weight is, the higher the proportion of the combination is. Because of the multi-head property and multidimensional output of the graph attention model compared with the conventional attention model, the weights from all head in different dimension are summed under different prediction length tasks in this paper, namely $A' \in \mathbb{R}^{T_p \times T_p}$. And the original heat map of weight matrix is shown in Fig. 11.a. In order to judge the importance of different combinations more intuitively, the original weight matrix values are mapped to the range of 0~1. Specifically, the softmax operation is carried out to control the sum of the importance of different combinations to be 1, and the processed heat map is shown in Fig. 11.b.

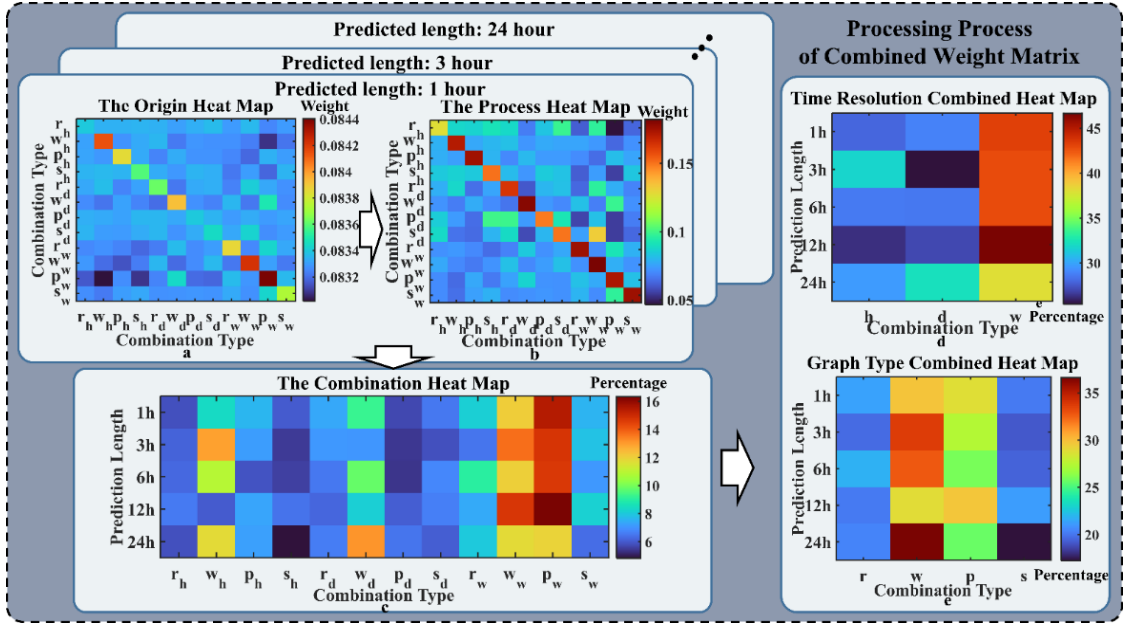

Fig. 11. Combined weight matrix processing process and result display

For the weight value $a_{ij} \in A'$ under any combination in the heat map, where $i = \{1, 2, ..., T_p\}$, $j = \{1, 2, ..., T_p\}$, the weight values represent the correlation degree among different combinations. As shown in Fig. 11.b, the autocorrelation degree of different combinations is the highest, while the correlation degree of combinations is lower.

In order to decouple the importance of different combinations under different prediction lengths, the influence distribution heat map of different combinations based on different prediction lengths is generated. According to the introduce of processed heat map, each column of heat map represents the correlation weight between each combination and combination of corresponding column. Based on summing the weight of each column in heat map, the total weight of combination of the corresponding column is calculated, and the total importance weight is represented as the corresponding combination. Specifically, corresponding to different prediction lengths, the processed heat map is summed by row, and then normalized and performed by operation of softmax to make the sum of importance weights of different combinations as 1, as shown in Fig. 11.c. Different combinations correspond to different importance degrees. The numbers in the figure represent the proportion of importance degrees, and the unit is percentage. Among the different combinations, the combination of hourly resolution and weighted topology map, daily input and weighted topology graph, weekly input and weighted topology graph, weekly input and traffic pattern graph have a large proportion under different prediction lengths.

Based on the above heat maps of different combined influence distribution under different prediction lengths, the type of graph and the resolution of input in the graph convolution are further decoupled, that is, the influence of spatial and temporal dimensions is decoupled respectively. Specifically, for the time resolution combined heat map, the combinations with the same resolution in all combinations are fused, and then the weights of spatial combinations are conducted by normalization and softmax operation. The data visualization results are shown in Fig. 11.d; For the combined heat map of graph type, the combinations of graph convolution based on the same graph in all combinations are integrated, and then the weights of temporal combinations are conducted by normalization and softmax operation. The data visualization results are shown in Fig. 11.e.

Based on the Fig. 11.e, with the increase of prediction length, the importance of daily resolution gradually increases and the importance of hourly resolution gradually decreases. For the comparison of different graphs, under different prediction lengths, the weighted topology graph is generally the most important, while the importance of traffic pattern graph is slightly lower than that of weighted topology graph, followed by topology graph and road attribute graph.

*5.3 Experiment 2: Comparison with Baseline*

To evaluate the predictive performance of MGCN-Attn, the model is compared with the following baseline. In these baselines, the inputs of SVC, CNN, LSTM, DCRNN and ST-GCN are the average road speed or flow. All baselines are optimized to output the best performance.

SVC: traffic prediction model based on support vector regression algorithm is trained. Here, the radial basis kernel is used, and the regularization parameter is 9.0.

CNN: a classical convolutional neural network for European spatial data modeling. Here, the network architecture by stacking volume layer and full connection layer is realized. And the convolution kernel size of convolution layer is [10,5], the step size is [5,4], the padding is [1,1], and the number of output channels is 234.

LSTM: a classical recurrent neural network for time series data modeling. Here, the time dimension of traffic data through the full connection layer is first linearly mapped, then the output of fully connection layer is inputted into the double-layer LSTM layer, and the final result through the two-layer full connection layer is outputted. The hidden layer dimension of the first full connection layer is 64, the hidden layer dimension of the double-layer LSTM is 64, the hidden layer dimension of the final two full connection layers is 32, and the output layer dimension is 5.

DCRNN: it refers to the method of [35], and the model predicts traffic by modeling the diffusion process of traffic flow on the graph. Its architecture includes GCN module, in which three diffusion convolution layers and one full connection layer are stacked. The first two layers of convolution increase the number of feature channels to 128, the last layer of convolution reduces the number of channels to 64, and the order of each layer of diffusion convolution is 3.

ST-GCN: it refers to the method of [26]. It combines graph and gated time convolution to simultaneously learn spatiotemporal patterns from traffic sequence data based on graph structure. The structure is stacked with two layers of sub spatial temporal graph convolution modules. Each module is composed of one layer of time convolution module, one layer of spatial map convolution module and one layer of time convolution module. Among them, the first layer time convolution module increases the number of feature channels to 64, the spatial map convolution module reduces the number of feature channels to 32, the last layer time convolution module increases the number of feature channels to 64, and the dimension of the hidden layer of the last full connection layer is 80.

The prediction accuracy and prediction consistency of DL-STFEE model and baselines in traffic grade prediction task on PeMS dataset are shown in Fig.12 respectively, and the specific values are shown in Table 4. the following important observations is drawn, compared with different baselines, DL-STFEE always maintains the optimal prediction accuracy and consistency. On average, the accuracy is improved by 5% and the consistency is improved by 3%. With the increase of prediction length, the fluctuation of prediction effect is very small, specifically less than 1%. Therefore, the model of DL-STFEE has greater advantages in long-term prediction.

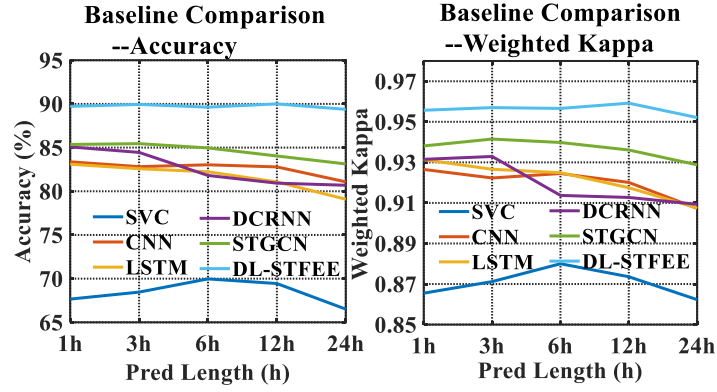

Fig. 12. Comparison diagram of DL-STFEE and baseline (for accuracy and weighted kappa coefficient)

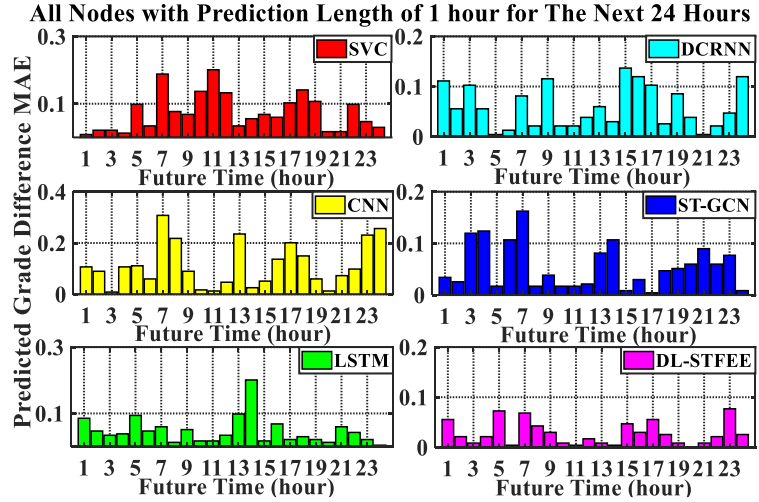

Fig. 13. The results of MAE of traffic grade prediction accuracy in one day by different methods

Secondly, for different baselines, the deep learning methods are significantly better than the machine learning method, and the accuracy is improved by about 15% on average. In different deep learning baselines, the model based on graph convolution neural network can achieve an accuracy of about 2% better than convolution neural network and recurrent neural network under appropriate graph definition methods. Different graph definition methods have a great impact on the spatial feature capture ability of graph convolution neural network, even about 3%. Finally, the predictive performance between convolution neural network and recurrent neural network in prediction effect is little, and the difference accuracy is less than 3%.

Then, in order to compare the prediction effects of DL-STFEE and different baselines more intuitively, the traffic grade's MAE (Mean Absolute Error) of different methods is chosen with data distributed in the 29th day as the evaluation index at the prediction length of 1 hour. The calculation method is shown in Fig. 13. Based on the prediction grade sequence $x^i$ $y^i$ of all roads in the whole graph at time point $i$, the prediction grade's MAE is calculated at this time point, and the above operation is repeated for all time points in a day. The smaller the MAE value at each time point is, the better the prediction effect of the whole graph is. The following figures show the comparison results with the predicted length of the next 1 hour. As shown in Fig. 13, the model DL-STFEE proposed in this paper shows better prediction effect at different time points than the baselines.

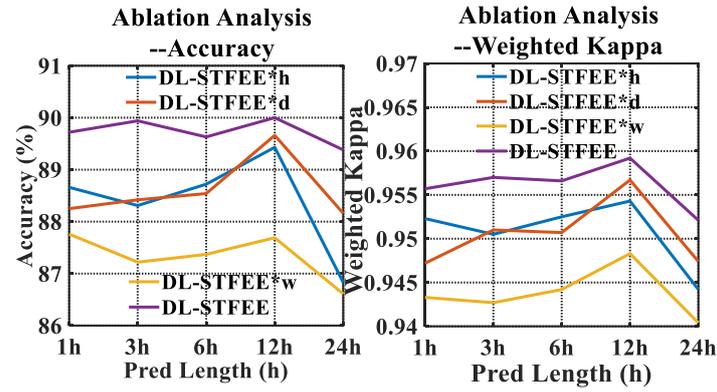

Fig. 14. The results of ablation analysis and real grade sequence

### 5.4 Experiment 3: Ablation Analysis

In order to study the effects of different resolutions of input in the proposed DL-STFEE model, the ablation research by comparing DL-STFEE and its variants are carried out. Among them, DL-STFEE*h, DL-STFEE*d and DL-STFEE*w are models with only hourly, daily and weekly single resolution input respectively.

Fig. 14. and Table 5 show the accuracy and consistency of DL-STFEE and its variants under the historical length of 24 hours and different prediction lengths. Among them, the multi-resolution input and single resolution input models maintain a high prediction accuracy and consistency, that is, the grade prediction accuracy is higher than 86%, and the weighted kappa coefficient is higher than 0.94. The model with three resolution inputs is better than the model with single resolution input in terms of accuracy and prediction consistency, and the accuracy is improved by about 2% on average. The prediction effect of the model based on hourly resolution input and daily resolution input is better than that of the model based on weekly resolution input, and the accuracy is improved by about 1% on average, while the prediction effect of the model based on hourly resolution input is not different from that of the model based on daily resolution input, and the accuracy difference is within 1%.

Table 4 Comparison table of DL-STFEE with baseline

| Prediction Length | 1h | | 3h | | 6h | | 12h | | 24h | |
|---|---|---|---|---|---|---|---|---|---|---|
| Type | ACC | Kappa | ACC | Kappa | ACC | Kappa | ACC | Kappa | ACC | Kappa |
| SVC | 0.676 | 0.865 | 0.684 | 0.871 | 0.699 | 0.880 | 0.694 | 0.873 | 0.665 | 0.862 |
| CNN | 0.83 | 0.927 | 0.828 | 0.922 | 0.83 | 0.925 | 0.828 | 0.92 | 0.811 | 0.907 |
| LSTM | 0.83 | 0.931 | 0.826 | 0.927 | 0.822 | 0.925 | 0.811 | 0.918 | 0.791 | 0.908 |
| DCRNN | 0.85 | 0.932 | 0.845 | 0.933 | 0.818 | 0.914 | 0.809 | 0.913 | 0.807 | 0.909 |
| ST-GCN | 0.85 | 0.938 | 0.855 | 0.942 | 0.85 | 0.94 | 0.84 | 0.936 | 0.831 | 0.929 |
| DL-STFEE | 0.9 | 0.956 | 0.899 | 0.957 | 0.896 | 0.957 | 0.9 | 0.959 | 0.894 | 0.952 |

Table 5 Comparison table of DL-STFEE with baseline

| Prediction Length | 1h | | 3h | | 6h | | 12h | | 24h | |
|---|---|---|---|---|---|---|---|---|---|---|
| Type | ACC | Kappa | ACC | Kappa | ACC | Kappa | ACC | Kappa | ACC | Kappa |
| DL-STFEE*h | 0.89 | 0.952 | 0.883 | 0.951 | 0.887 | 0.953 | 0.894 | 0.954 | 0.868 | 0.944 |
| DL-STFEE *d | 0.88 | 0.947 | 0.884 | 0.951 | 0.885 | 0.951 | 0.897 | 0.957 | 0.882 | 0.947 |
| DL-STFEE *w | 0.88 | 0.943 | 0.872 | 0.943 | 0.874 | 0.944 | 0.877 | 0.948 | 0.866 | 0.94 |
| DL-STFEE | 0.9 | 0.956 | 0.899 | 0.957 | 0.896 | 0.957 | 0.9 | 0.959 | 0.894 | 0.952 |

## VI. Discussion

The above experimental parts are tested from three perspectives: the impact of different combinations, comparison with baseline and ablation analysis. The internal deep information of the results plays a guiding role in the task of traffic grade prediction. The laws in detail based on these three perspectives are analyzed in the following paper:

1. According to the influence of different combinations, the hourly resolution data has strong guiding significance for short-term prediction (in the next few hours), namely it has short periodicity. Daily resolution data has strong effect for long-term prediction (in the next day), i.e., it has long periodicity. And weekly resolution data has general guiding significance for both short-term prediction and long-term prediction. The result is very consistent with our experience.

2. According to the influence of different graph network definition methods, the topological relationship and traffic pattern similarity relationship between roads can better reflect the spatial similarity of traffic data. For the traffic road network, the topological relationship directly reflects the spatial distribution of traffic, and the distance in different physical distances directly affects the spatial distribution similarity of road traffic data; The similarity relationship of traffic patterns is defined based on the historical traffic information of the roads. The roads with similar historical traffic states are relatively more similar in the future traffic distribution, that is, the similarity between different patterns indirectly affects the similarity of road spatial distribution; The similarity relationship of road attributes determines the similarity of traffic spatial distribution between roads based on the maximum speed and length of roads. If the topological relationship directly affects the spatial state distribution of traffic condition based on the fixed spatial distribution, and the traffic pattern relationship indirectly affects the spatial state distribution of traffic condition through the time-varying historical pattern, the similarity relationship of road attributes is time-fixed and indirectly affects the similarity relationship of spatial state. Compared with the first two relationships, the similarity of road attributes has less impact on the similarity of traffic state is also very reasonable.

3. According to the comparison of prediction effects between different models, some rules are found. Deep learning model is obviously better than machine learning in traffic prediction effect, because deep learning can realize the modeling of temporal and spatial correlation of traffic data, while machine learning cannot capture the temporal or spatial correlation of traffic data because of its simple structure. The model based on graph convolution neural network realizes the modeling of traffic spatial-temporal correlation characteristics at the same time. Compared with the convolution neural network model with single modeling spatial correlation and the recurrent neural network model with single modeling time correlation, the graph convolution neural network has better feature learning effect; Compared with other spatial-temporal correlation modeling models for traffic data, DL-STFEE has the characteristics of combining multi-resolution input and multi traffic graph definition. Through the extraction and fusion of spatial-temporal features, DL-STFEE achieves better prediction effect and also has the functions of decoupling and explicit analysis of spatiotemporal feature combinations in the combination fusion layer, and it provides some guidance for future traffic prediction tasks.

4. For the analysis of ablation experiment, on the one hand, different sub models have good prediction results, and sub models show the contribution of prediction effect among the three types of time resolution, that is, inputting the three types of time resolution is meaningful for traffic prediction tasks. On the other hand, the comparison of prediction performance among sub models shows that the effect of model with hourly resolution and daily resolution input is better than that with weekly resolution input. With the increase of prediction length, the prediction effect gradually changes from the relatively optimal hourly resolution group to the optimal daily resolution group, the conclusion from it is the data of historical hours can better reflect the traffic conditions in short-term prediction, and the data of several days in history can better reflect the periodicity of traffic conditions in long-term prediction. Finally, the prediction effect of the model based on weekly resolution input is more stable. The input of weekly resolution can further improve the prediction effect of the other two resolution groups, indicating that it can better reflect the periodicity of traffic data. The analysis of the ablation experiment is consistent with the above analysis of the heat map at different resolutions, and the analysis further verifies the property of interpretable of the heat map.

Based on the rules summarized, the traffic prediction task is completed with more instructive. However, the current model still has some shortcomings.

1. The model can only realize the single step prediction with different prediction lengths, and still cannot complete the grade prediction task for a continuous period of time in the future.

2. The data in this paper are based on the data collected by traffic road sensors alone, and fail to consider the weather, events and other factors in the city.

Therefore, the multi-step prediction and considering the external factors such as weather and events at the same time are the direction of future improvement. For the multi-step prediction, the seq2seq architecture is considered to complete the better prediction of the future multi-step prediction task. For extracting useful traffic features from external factors, the external factor features extraction module based on CNN model to extract traffic features is considered in the future.

## VII. CONCLUSION

In this paper, the model of DL-STFEE is proposed, and it's a framework based on graph convolution and attention mechanism to model and predict complex traffic data. The model extracts the spatiotemporal correlation features of the traffic data through three kinds of temporal resolutions and four graph matrices, and measures the importance of different spatiotemporal feature combinations through the multi-dimensional graph attention mechanism. The large number of experiments on large-scale traffic data sets are conducted, and the results show that the model structured in this paper has advantages in traffic condition grade prediction compared with the existing models.

Although the model DL-STFEE has achieved promising results, it still has a certain optimization space. Specifically, the traffic prediction considering the environmental factors and the multi-step traffic prediction task form a hot topic. Integrating these improvements together may provide a powerful model to predict the traffic state with a higher performance.

<div align="center">REFERENCES</div>